\documentclass{article}

\usepackage[final,main]{neurips_2026}

\usepackage[utf8]{inputenc} 
\usepackage[T1]{fontenc}    
\usepackage{hyperref}       
\usepackage{url}            
\usepackage{booktabs}       
\usepackage{amsfonts}       
\usepackage{nicefrac}       
\usepackage{microtype}      
\usepackage[table]{xcolor}

\usepackage{amsmath}  
\usepackage{amssymb} 
\usepackage{graphicx}

\usepackage[most]{tcolorbox}
\usepackage{listings}
\usepackage{verbatim}
\usepackage{caption}
\usepackage{wrapfig}
\usepackage{arydshln}
\usepackage{multirow}
\definecolor{neoncyan}{RGB}{0, 230, 240}

\usepackage{booktabs}
\usepackage{pifont} 

\newtcolorbox{instructionsbox}[1][]{
  colback=white!90!cyan!10,
  colframe=neoncyan!90!gray,
  title=#1,
  boxrule=0.6pt,
  arc=2pt,
  left=6pt, right=6pt, top=6pt, bottom=6pt
}

\newtcolorbox{errorbox}[1][]{
  colback=pink!5,        
  colframe=pink!60!red!80,
  title=#1,
  boxrule=0.6pt,
  arc=2pt,
  left=6pt, right=6pt, top=6pt, bottom=6pt
}

\usepackage{adjustbox}  
\usepackage{caption}
\title{Retrieval as Exploration: \\ Retriever-Free Retrieval via Corpus-Traversing MCTS}

\author{%
  Chaeeun Kim \\
  Grounded \\
  \texttt{chaeeun@groundedin.io} \\
  \And
  Seungone Kim \\
  Carnegie Mellon University \\
  \texttt{seungone@cmu.edu} \\
}

\begin{document}

\maketitle
\begingroup
\renewcommand{\thefootnote}{}
\footnotetext[0]{Correspondence to Chaeeun Kim <chaeeun@groundedin.io>.}
\endgroup

\begin{abstract}
Large Reasoning Models (LRMs) have demonstrated remarkable capabilities in multi-step reasoning and calling search tools at appropriate steps. However, existing retrieval-augmented reasoning approaches still rely on separate retrieval models, limiting LRMs to deciding \textit{when} to retrieve and \textit{how} to formulate queries rather than leveraging their pretrained knowledge and reasoning capabilities for retrieval itself. This structural separation introduces a representation bottleneck, as retrieval relies on separately learned representations that may obscure fine-grained distinctions needed for reasoning. Moreover, these approaches commonly employ fixed top-k retrieval, which can introduce superfluous documents and increase retrieval noise and context overhead. To address this, we reformulate retrieval from query-document relevance matching into the exploration of answer-containing paths within the corpus, and propose \textbf{FREESON} (Retriever-\textbf{FREE} Retrieval-Augmented Rea\textbf{SON}ing), a framework that enables LRMs to directly navigate the original corpus text to locate information that satisfies their specific needs, serving as both the reasoner and the retriever. Specifically, we introduce \textbf{CT-MCTS} (\textbf{C}orpus-\textbf{T}raversing \textbf{M}onte \textbf{C}arlo \textbf{T}ree \textbf{S}earch), a retrieval-specialized variant of Monte Carlo Tree Search (MCTS). CT-MCTS iteratively expands and evaluates exploration paths, starting from multiple promising entry points across the corpus. This allows LRMs to selectively reference specific segments identified during traversal, instead of fetching a fixed top-k set. Across nine benchmarks, FREESON improves average EM and F1 by 15 points over four retriever-based baselines on five general-domain QA tasks, while matching the strongest baseline. It also achieves a 12-point gain on fact-checking and a 13-point gain on medical QA under zero-shot transfer, approaching the performance of the much larger medical LM MedGemma-27B.



\end{abstract}

\section{Introduction}


Recent advances in large reasoning models have enabled agentic reasoning frameworks in which models interact with external tools and environments throughout the reasoning process. For knowledge-intensive tasks, these frameworks often interleave multi-step reasoning with external knowledge retrieval to acquire up-to-date or missing information on demand, making retrieval quality critical for guiding subsequent reasoning steps~\citep{jiang2023activeretrievalaugmentedgeneration, press2023measuringnarrowingcompositionalitygap, asai2023selfraglearningretrievegenerate, li2025searcho1agenticsearchenhancedlarge, jin2025searchr1trainingllmsreason, 
song2025r1searcherincentivizingsearchcapability, yao2023reactsynergizingreasoningacting, 
wang2025rareretrievalaugmentedreasoningmodeling, schick2023toolformerlanguagemodelsteach}. As reasoning proceeds step by step, each retrieval often targets a specific piece of knowledge, requiring not only semantic relevance but also precise lexical matching to identify the exact information needed. However, conventional retrieval approaches rely on separate retrieval models, typically based on dual-encoder architectures, whose embedding-based relevance matching often fails to reflect the precise requirements by the reasoning model, retrieving documents that are broadly relevant but miss the exact details needed~\citep{wang2023neuralcorpusindexerdocument, kim2024exploringpracticalitygenerativeretrieval, magesh2024hallucinationfreeassessingreliabilityleading}. For example, given the query ``Where was the place of burial of John Tuchet, 6th Baron Audley's father?'', E5$_{\text{base}}$~\citep{wang2024textembeddingsweaklysupervisedcontrastive}, a widely used dense retriever, assigns higher similarity scores to ``John Tuchet, 8th Baron Audley'' and ``George Tuchet, 9th Baron Audley'' (both incorrect) than to ``John Tuchet, 6th Baron Audley'' (correct). Unable to distinguish documents that differ by \textit{only a single character}, the retriever returns an incorrect document in the first hop and consequently fails to reach the final answer document, ``James Tuchet, 5th Baron Audley''. Moreover, since these retrievers only rank documents by relative similarity without determining which ones actually contain the required evidence, they are typically used with a fixed top-k cutoff, often including extraneous documents that add noise and consume the limited context window~\citep{shi2023largelanguagemodelseasily, liu2023lostmiddlelanguagemodels, jin2024longcontextllmsmeetrag, li2025searcho1agenticsearchenhancedlarge}.

To address this issue, prior works have proposed better representation learning methods, or scaling up either the model size or the amount of training data to enhance retrieval performance~\citep{izacard2022unsuperviseddenseinformationretrieval, ram2022learningretrievepassagessupervision, wang2024textembeddingsweaklysupervisedcontrastive, wang2024improvingtextembeddingslarge, lee2024geckoversatiletextembeddings, shao2025reasonirtrainingretrieversreasoning}. However, maintaining separate reasoner and retriever models introduces additional training, deployment, and maintenance overhead, including the cost of building and managing large embedding-based indexes~\citep{zhang2024onegenefficientonepassunified, reichman-heck-2024-dense, kim2024exploringpracticalitygenerativeretrieval}. In this paper, we revisit the conventional retrieval-augmented reasoning paradigm and pose a question: \textit{Can a reasoner autonomously navigate a vast corpus by leveraging its inherent discriminative power to acquire the necessary knowledge?}

To address this question, we shift our focus from embedding-based relevance matching to locating answer-containing paths within the corpus and propose a novel framework called \textbf{FREESON} (Retriever-\textbf{FREE} Retrieval-Augmented Rea\textbf{SON}ing). In this framework, a single LRM serves as both the reasoner and the retriever, autonomously navigating the original corpus text to acquire relevant evidence. To implement this, we first introduce \textbf{CT-MCTS} (\textbf{C}orpus-\textbf{T}raversing \textbf{M}onte \textbf{C}arlo \textbf{T}ree \textbf{S}earch), a retrieval-specialized MCTS~\citep{10.1007/11871842_29, Silver2016MasteringTG, chen2024alphamathzeroprocesssupervision, liu2024dontthrowawayvalue, zhang2024restmctsllmselftrainingprocess}
, in which the LRM itself directly generates multiple promising entry nodes across the corpus and explores answer-bearing locations by expanding and evaluating each path with its own discriminative capability, while the search space is constrained to token sequences that appear in the corpus (\S~\ref{method:ct_mcts}). 


Implementing such a retrieval-oriented MCTS introduces two key challenges:
(1) the search operates at an extremely fine granularity, with token-level nodes, so each node carries too little semantic content to be meaningfully evaluated; and 
(2) accurate node value estimates are required to guide the search toward answer-bearing regions, yet it remains unclear which value estimation strategies are effective for corpus traversal.
To address the first challenge, we extend the node unit from single tokens to multi-token spans, preserving semantic richness. We further incorporate stochastic beam search into the node expansion, allowing the LRM to generate multiple promising continuations in parallel. For the second challenge, we investigate self-evaluation, off-policy value learning using LLM-generated synthetic trajectories, and on-policy value learning using CT-MCTS rollouts. We find that on-policy learning consistently outperforms off-policy learning, while self-evaluation achieves comparable performance, allowing our retriever-free framework to be entirely training-free (\S~\ref{anlaysis:reward_model_comparison}).


We evaluate \textsc{FREESON} on nine benchmarks covering single- and multi-hop QA, fact-checking, and medical QA. On five open-domain QA benchmarks, FREESON improves EM and F1 by 14.9 points on average over four retriever-based reasoning models, performing on par with the strongest baseline, Search-R1, without an external retriever. On FEVER, it achieves the highest accuracy with a 12-point gain over Search-R1. In zero-shot medical QA, FREESON-7B achieves competitive performance with MedGemma-27B and outperforms baselines using the medical retriever MedCPT (\S~\ref{section:main_results}). Our retrieval-specialized \textsc{CT-MCTS} plays a key role in this performance. Flexible node granularity improves EM and F1 by 27.5 points on average over single-token nodes (\S~\ref{analysis:node_granularity}), and multi-node expansion guided by the LRM improves performance by 13 points over single-node expansion (\S~\ref{analysis:multiple_expansion}). \textsc{FREESON} does not require any training and is applicable to arbitrary LMs provided their output logits are accessible. Particularly, it is well suited for domain-specific applications where additional training is costly or impractical, as it directly explores and reasons over the corpus at inference time.

\section{Related Works}

\subsection{Retrieval-Augmented Generation}
Large Language Models (LLMs) achieve strong performance in language understanding and generation but suffer from hallucinations in domain-specific tasks. Retrieval-Augmented Generation (RAG) alleviates this by retrieving relevant documents before generation, improving factuality~\citep{guu2020realmretrievalaugmentedlanguagemodel, lewis2021retrievalaugmentedgenerationknowledgeintensivenlp, borgeaud2022improvinglanguagemodelsretrieving, izacard2022atlasfewshotlearningretrieval}. Early approaches follow a retrieve-then-generate framework~\citep{lewis2021retrievalaugmentedgenerationknowledgeintensivenlp}, while later work studied what~\citep{khandelwal2020generalizationmemorizationnearestneighbor, borgeaud2022improvinglanguagemodelsretrieving}, how~\citep{ram2023incontextretrievalaugmentedlanguagemodels}, and when to retrieve~\citep{guu2020realmretrievalaugmentedlanguagemodel}.

Retrieval has increasingly been integrated into reasoning frameworks. ReAct~\citep{yao2023reactsynergizingreasoningacting} combines reasoning with external tool use, Self-Ask~\citep{press2023measuringnarrowingcompositionalitygap} generates sub-questions, FLARE~\citep{jiang2023activeretrievalaugmentedgeneration} uses look-ahead queries, and Self-RAG~\citep{asai2023selfraglearningretrievegenerate} learns when to retrieve. More recent methods, such as Search-o1~\citep{li2025searcho1agenticsearchenhancedlarge} and Search-R1~\citep{jin2025searchr1trainingllmsreason}, further integrate retrieval into multi-step reasoning with Large Reasoning Models (LRMs)~\citep{deepseekai2025deepseekr1incentivizingreasoningcapability, zhong2024evaluationopenaio1opportunities, qwen2025qwen25technicalreport}. However, these approaches still largely rely on separate retrievers, particularly embedding-based retrieval, which introduces a representation bottleneck that may obscure fine-grained distinctions needed for reasoning. In contrast, \textsc{FREESON} reformulates retrieval as corpus exploration, enabling LRMs to directly navigate the original corpus text to pinpoint grounding evidence without a separate retriever.


\subsection{LLMs with Monte Carlo Tree Search}

Monte Carlo Tree Search (MCTS)~\citep{10.1007/11871842_29, wei2025unifyingtreesearchalgorithm} has recently emerged as a powerful framework for enhancing the multi-step reasoning capabilities of LLMs. By structuring the reasoning process as a search tree consisting of selection, expansion, simulation, and backpropagation, MCTS allows models to explore diverse solution paths and evaluate them through value-based guidance. In particular, AlphaMath~\citep{chen2024alphamathzeroprocesssupervision} applies MCTS for math reasoning with value models, ReST-MCTS*~\citep{zhang2024restmctsllmselftrainingprocess} leverages process reward guidance for self-training, and PPO-MCTS~\citep{liu2024dontthrowawayvalue} integrates PPO value networks with MCTS at inference. To the best of our knowledge, \textsc{FREESON} is the first to adapt MCTS specifically for the internal mechanics of retrieval. Unlike prior works that use MCTS only to plan high-level reasoning steps, our Corpus-Traversing MCTS (\textsc{CT-MCTS}) enables a single LRM to directly traverse the raw text corpus to locate answer-containing paths.

\subsection{Generative Information Retrieval}

Generative Information Retrieval (GenIR)~\citep{decao2021autoregressiveentityretrieval, tay2022transformermemorydifferentiablesearch, wang2023neuralcorpusindexerdocument, zeng2023scalableeffectivegenerativeinformation, zeng2024planningaheadgenerativeretrieval, bevilacqua2022autoregressivesearchenginesgenerating, li2023multiviewidentifiersenhancedgenerative, li2023learningrankgenerativeretrieval, li2025matchinggenerationsurveygenerative, mehta2023dsiupdatingtransformermemory, Chen_2022, kim2024exploringpracticalitygenerativeretrieval, tang2024selfretrievalendtoendinformationretrieval, kim2025legalsearchlmrethinkinglegalcase} integrates autoregressive LMs into retrieval, focusing on internalizing corpus information directly into the parameters. The generative retrieval model is trained to map queries to specific document identifiers, such as titles \citep{decao2021autoregressiveentityretrieval}, document IDs ~\citep{tay2022transformermemorydifferentiablesearch, wang2023neuralcorpusindexerdocument, zeng2023scalableeffectivegenerativeinformation, zeng2024planningaheadgenerativeretrieval}, or specific text spans \citep{bevilacqua2022autoregressivesearchenginesgenerating, li2023multiviewidentifiersenhancedgenerative, li2023learningrankgenerativeretrieval, kim2025legalsearchlmrethinkinglegalcase}. Consequently, the performance of GenIR heavily depends on the model's memorization capacity, requiring extensive pre-training or fine-tuning. In contrast, \textsc{FREESON} eliminates the need for such internalized memorization. Instead of recalling fixed identifiers, the reasoner directly locates grounding evidence by traversing a global text index in real-time. This architectural shift ensures high scalability and zero-shot adaptability to dynamic data.


\section{Methodology}
\begin{figure}[t!]
{
\centering
    \centering
    \includegraphics[width=\textwidth]
    {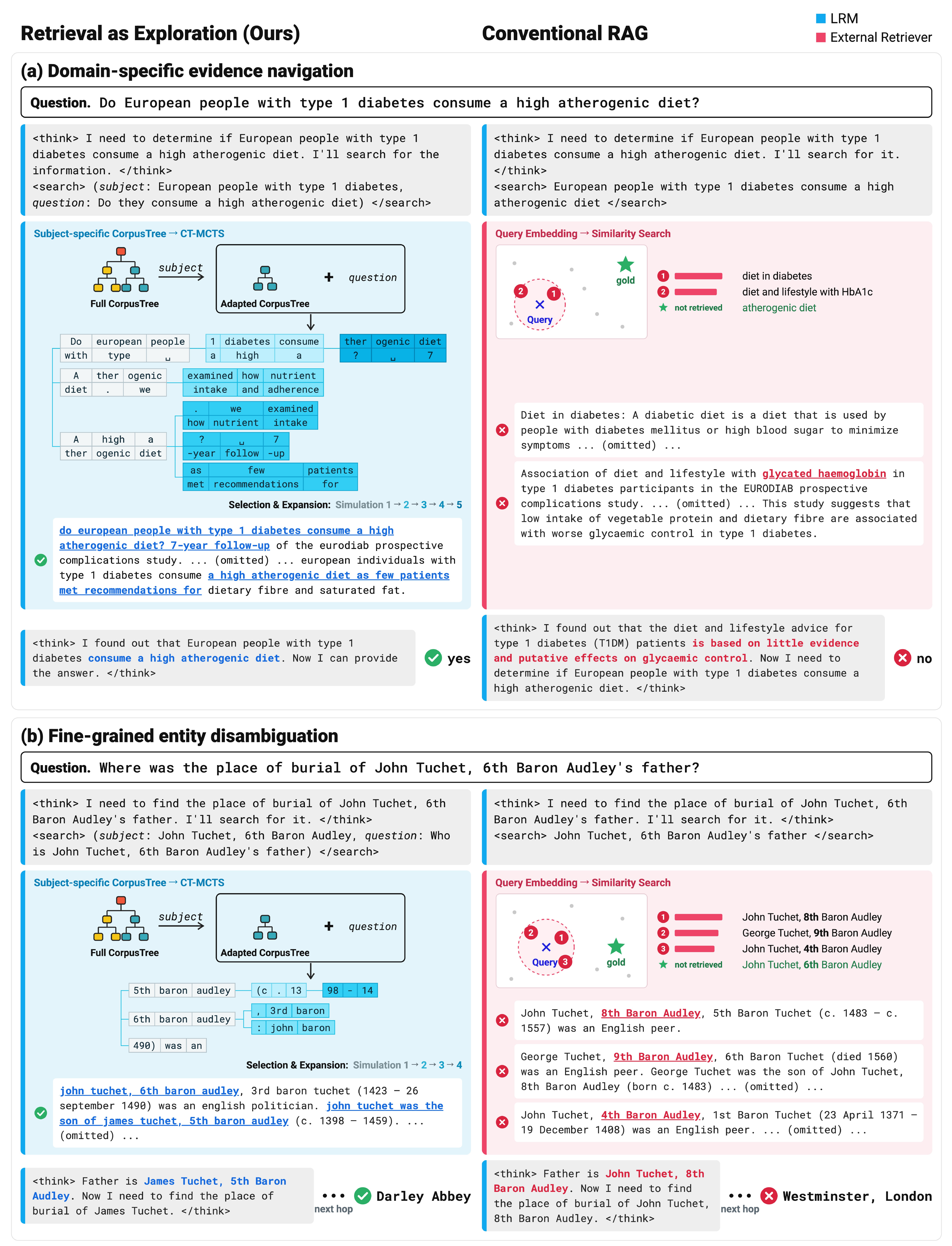}
    \caption{Overview of retrieval-augmented reasoning on (a) domain-specific evidence navigation and (b) fine-grained entity disambiguation. FREESON (left) runs CT-MCTS over a subject-adapted CorpusTree, expanding nodes of $G$ tokens ($G=6$ in (a) and $G=3$ in (b); colored by simulation order), with each simulation scored by a rollout (not shown). Search-R1 (right) retrieves related but non-answering documents using a dense retriever. Traces and retrieved passages are verbatim model outputs.
}

    \label{figure:main_figure}
}
\end{figure}

In this section, we detail how \textsc{FREESON} autonomously explores the original corpus text to acquire the necessary knowledge. See Figure \ref{figure:main_figure} for an illustration of this process.

\subsection{Dynamically Adapted Search Space}
When search is required, \textsc{FREESON} first generates a query in the form of \textit{(subject: [subject], question: [question])}, where [subject] is the main entity or phrase, and [question] is the corresponding question. Based on the subject, we adaptively narrow the search space within the \textit{CorpusTree}, which is implemented using an FM-index, with token-level n-gram backoff. The resulting \textit{Adapted CorpusTree} filters out irrelevant documents, similar to how ANN search prunes candidates in embedding retrieval. We use a few-shot prompt to elicit how information might be expressed in the corpus (Appendix~\ref{appendix:r2r_retrieval_prompt}).

\subsection{CT-MCTS for Direct Corpus Exploration}
\label{method:ct_mcts}
Given a question and a subject-adapted CorpusTree, \textsc{FREESON} locates the appropriate information using \textsc{CT-MCTS} designed to reinforce paths that are likely to contain the correct answer. \textsc{CT-MCTS} operates in a token-level search space where the LLM's probability distribution is dynamically masked by the CorpusTree, allowing only valid sequences found in the corpus. 
The CorpusTree is implemented using an FM-Index~\citep{10.5555/795666.796543, bevilacqua2022autoregressivesearchenginesgenerating}, a compressed full-text index based on the Burrows-Wheeler Transform (BWT)~\citep{10.1145/382780.382782} that enables efficient substring search through suffix-array-based indexing. This allows CT-MCTS to efficiently navigate token sequences constrained to the original corpus. In the following, we describe the key components of \textsc{CT-MCTS}.

\paragraph{Selection.}
We initialize the search tree by expanding an empty root into multiple promising entry points across the corpus. We then employ the widely used UCT selection function~\citep{10.1007/11871842_29, 3f17dd9bdc65470ca9b38ac58ba954ef} to recursively select, at each node \(s\), the child node \( a \in \text{CorpusTree}(s) \) with the highest UCT score until reaching a leaf node.


\[
a^* = \arg\max_a \left[ Q(s, a) + \lambda \cdot \sqrt{\frac{\log \sum_b N(s, b)}{1 + N(s, a)}} \right]
\]

Here,  \( Q(s, a) \) is the average value of taking action $a$ from node $s$,  \( N(s, a) \) is the number of times action a has been selected from node $s$, and \( \lambda \) is a scalar balancing exploration and exploitation. The action space is constrained by a CorpusTree according to the expansion process described below.

\paragraph{Granularity-aware multi-node expansion.} The second step is to expand the selected node by determining the promising next search directions. Our expansion approach differs from conventional MCTS in two key ways. 

(1) \textit{Granularity-aware node expansion}: Each node in our search tree contains a sequence of tokens (length $G$), rather than a single token (e.g., ``5th Baron Audley'' instead of ``5th''; see Figure~\ref{figure:main_figure}).  This allows the model to make more context-aware and semantically meaningful decisions at each step, while also enabling faster search by traversing multiple steps in a single move. This allows the reasoning model to contribute more actively to the expansion process. Through this, we strengthen the effectiveness of CorpusTree-guided search while still adhering to token-level constraints for retrieval (\S~\ref{analysis:node_granularity}).

(2) \textit{Multi-node expansion per simulation}: Rather than expanding a single node per simulation, we expand the $M$ candidate children based on the LRM's next-token probabilities (e.g., ``, 3rd baron'' and ``: john baron'' are expanded in a single simulation; see Figure~\ref{figure:main_figure}). This allows the model to better utilize the LRM's outputs during expansion, resulting in substantial performance gains (\S~\ref{analysis:multiple_expansion}).

To enable the two features, we employ stochastic beam search decoding. Let $\mathcal{A}_s = \text{CorpusTree}(s)$ denote the set of valid next tokens for the current selected path $s$, and $\tilde{\mathcal{A}}_s = \{ \tilde{a}_1, \dots, \tilde{a}_k \} \subset \mathcal{A}_s$ be the top-$k$ tokens ranked by log-probability. Final candidates are sampled from $\tilde{\mathcal{A}}_s$ using multinomial sampling. For each candidate, we iteratively extend the sequence by sampling tokens until either a predefined per-node token limit $G$ is reached or no valid tokens remain ($\mathcal{A}_s = \emptyset$), while maintaining the top-$M$ paths ranked by cumulative log-probability at each step.

\paragraph{Evaluating answer presence in search trajectories.}
For each newly expanded node, we perform a rollout to estimate whether the resulting path leads toward an answer-bearing region. The rollout follows a greedy decoding process, guided by CorpusTree at each step, and continues until either the maximum sequence length is reached or no valid tokens remain.

Once the rollout terminates, the resulting path \(s_{\text{final}}\) is evaluated using an answer presence-aware value estimation strategy, which takes as input the question and \(s_{\text{final}}\) and outputs a value estimate \(\hat{y} \in [0,1]\). The resulting value estimate \(\hat{y}\) is backpropagated to update \(Q(s,a)\), while the corresponding visit count \(N(s,a)\) is incremented. We describe the value estimation strategies in Section~\ref{sec:value_estimation}.

\paragraph{Document selection.} After CT-MCTS exploration, we consider three levels of contextual information: exact spans (Direct Path), surrounding context windows (Window Expansion), and complete source documents (Complete Document). All three strategies retrieve information based on the explored paths without relying on a fixed top-k document cutoff. We adopt Complete Document to preserve contextual information and avoid potential information loss.



\subsection{Value Estimation Strategies}
\label{sec:value_estimation}

We investigate training-free self-evaluation, on-policy learning, and off-policy learning for value estimation. Our results show that both self-evaluation and on-policy learning are effective, and we adopt on-policy estimation in FREESON (\S~\ref{analysis:value_estimation}). For the latter two, we train a classification head on the frozen reasoning model using binary cross-entropy loss. During training, we use an 80GB A100 GPU.

\paragraph{Training-free self-evaluation.} The reasoning model directly evaluates CT-MCTS rollouts without additional training (Appendix~\ref{appendix:self_evaluation_prompt}).

\paragraph{On-policy learning from CT-MCTS rollouts.} We collect approximately 15,000 rollout-label pairs from 400 PopQA~\citep{mallen2023trustlanguagemodelsinvestigating} examples. For each rollout, we prompt the reasoning model with the question and candidate path to generate an answer, assigning a soft label based on its agreement with the ground truth (1.0 for a full match, 0.8 for a partial match, and 0.0 otherwise; see Appendix~\ref{appendix:r2r_reward_prompt}).

\paragraph{Off-policy learning from synthetic trajectories.} Using GPT-4o and 2WikiMultihopQA~\citep{ho-etal-2020-constructing}, we generate 147,755 path-value pairs from queries and their evidence sentences, varying path length, relevance, and answer entailment (Appendix~\ref{appendix:value_network_synthetic}).

\section{Experiments}
\subsection{Benchmarks}

We evaluate \textsc{FREESON} on nine benchmarks covering fact-checking, general-domain QA, and medical QA. We use Wikipedia corpora with 6M passages for 2WikiMultihopQA and 21M DPR passages for the remaining general-domain tasks, and a medical textbook corpus of 125.8K snippets for medical QA. We report accuracy for fact-checking and medical QA, and EM and F1 for general-domain QA. Further details are provided in Appendix~\ref{appendix:stats}.

\textbf{Fact-checking.} \textsc{FEVER}~\citep{thorne-etal-2018-fever} evaluates claim verification with three labels: SUPPORTS, REFUTES, and NOT ENOUGH INFO.

\textbf{General-domain QA.} For multi-hop QA, we use \textsc{2WikiMultihopQA}~\citep{ho-etal-2020-constructing} for cross-page reasoning, \textsc{MuSiQue}~\citep{trivedi-etal-2022-musique} for 2–4 hop compositional reasoning, and \textsc{HotpotQA}~\citep{yang2018hotpotqadatasetdiverseexplainable} for multi-document reasoning with supporting facts. For single-hop QA, we use \textsc{PopQA}~\citep{mallen2023trustlanguagemodelsinvestigating}, which contains entity-centric factual questions, and \textsc{TriviaQA}~\citep{joshi2017triviaqalargescaledistantly}, which includes trivia questions with web evidence.

\textbf{Medical QA.} We use \textsc{MedQA-US}~\citep{jin2020diseasedoespatienthave} for USMLE-style questions, \textsc{PubMedQA}~\citep{jin2019pubmedqadatasetbiomedicalresearch} for biomedical research questions, and the yes/no subset of \textsc{BioASQ}~\citep{krithara2023bioasq}.

\subsection{Baselines}

We compare \textsc{FREESON} against three categories of baselines: \textbf{reasoning without retrieval}, which relies on parametric knowledge; \textbf{retrieve-then-reason}, which retrieves documents before reasoning; and \textbf{interleaved reasoning and retrieval with external retrievers}, which alternates reasoning and retrieval. Evaluation details and prompts are provided in Appendices~\ref{appendix:eval_details} and~\ref{appendix:prompt_template}.

\textbf{General-domain QA and fact-checking.} We evaluate four interleaved reasoning baselines with external retrievers: \textsc{FLARE}~\citep{jiang2023activeretrievalaugmentedgeneration}, which triggers retrieval based on token-level uncertainty; \textsc{Self-Ask}~\citep{press2023measuringnarrowingcompositionalitygap}, which decomposes questions into sub-questions; \textsc{Search-o1}~\citep{li2025searcho1agenticsearchenhancedlarge}, which incorporates a Reason-in-Documents module; and \textsc{Search-R1}~\citep{jin2025searchr1trainingllmsreason}, which trains retrieval-augmented reasoning through reinforcement learning. All baselines using external retrievers use Qwen2.5-7B as the backbone and E5$_{\text{base}}$~\citep{wang2024textembeddingsweaklysupervisedcontrastive} as the retriever. FREESON adopts the Search-R1 checkpoint (SearchR1-nq\_hotpotqa\_train-qwen2.5-7b-em-ppo) as its reasoning backbone, without an external retriever.

\begin{table*}[t!]
\centering
\fontsize{12}{16}\selectfont
\resizebox{\textwidth}{!}{
\setlength{\dashlinedash}{1.8pt}
\setlength{\dashlinegap}{2.2pt}
\begin{tabular}{lccccccccccc}
    \toprule
     & \multicolumn{1}{c}{\textbf{Fact-checking}} & \multicolumn{6}{c}{\textbf{Multi-hop QA}} & \multicolumn{4}{c}{\textbf{Single-hop QA}} \\
     \cmidrule(lr){2-2} \cmidrule(lr){3-8} \cmidrule(lr){9-12}
     \textbf{Method}
     & \multicolumn{1}{c}{\textbf{FEVER}}
     & \multicolumn{2}{c}{\textbf{2WikiMultihopQA}} & \multicolumn{2}{c}{\textbf{MuSiQue}} & \multicolumn{2}{c}{\textbf{HotpotQA}}
     & \multicolumn{2}{c}{\textbf{PopQA}} & \multicolumn{2}{c}{\textbf{TriviaQA}} \\
     \cmidrule(lr){2-2} \cmidrule(lr){3-4} \cmidrule(lr){5-6} \cmidrule(lr){7-8} \cmidrule(lr){9-10} \cmidrule(lr){11-12}
     & ACC & EM & F1 & EM & F1 & EM & F1 & EM & F1 & EM & F1 \\

    \midrule
    \multicolumn{12}{l}{\textbf{\textit{Reasoning without retrieval}}} \\
    GLM-4.7-Flash-30B-A3B & 0.66 & 0.29 & 0.35 & 0.09 & 0.18 & 0.27 & 0.37 & 0.30 & 0.35 & \textbf{0.64} & \textbf{0.73} \\
    Qwen3.8-27B & \underline{0.69} & 0.30 & 0.35 & 0.10 & 0.18 & 0.26 & 0.37 & 0.23 & 0.27 & 0.55 & 0.64 \\
    Gemma-4-12B & 0.61 & 0.24 & 0.28 & 0.06 & 0.14 & 0.23 & 0.33 & 0.19 & 0.22 & 0.53 & 0.62 \\
    R1-Distill-Qwen-7B & 0.38 & 0.18 & 0.20 & 0.01 & 0.03 & 0.11 & 0.15 & 0.07 & 0.10 & 0.13 & 0.17 \\
    \cdashline{1-12}
    Qwen2.5-7B & 0.33 & 0.19 & 0.23 & 0.02 & 0.06 & 0.13 & 0.19 & 0.09 & 0.11 & 0.26 & 0.31 \\

    \midrule
    \multicolumn{12}{l}{\textbf{\textit{Retrieve-then-reason}}} \\
    Qwen2.5-7B + E5 & 0.46 & 0.19 & 0.23 & 0.02 & 0.05 & 0.10 & 0.15 & 0.13 & 0.16 & 0.15 & 0.19 \\

    \midrule
    \multicolumn{12}{l}{\textbf{\textit{Interleaved reasoning and retrieval (with retriever)}}} \\
    \multicolumn{12}{l}{\textcolor{red}{\normalsize Backbone: Qwen2.5-7B}} \\
    FLARE-7B & 0.21 & 0.27 & 0.32 & 0.06 & 0.14 & 0.21 & 0.28 & 0.20 & 0.28 & 0.29 & 0.41 \\
    Self-Ask-7B & 0.59 & 0.22 & 0.28 & 0.03 & 0.09 & 0.18 & 0.27 & 0.21 & 0.24 & 0.33 & 0.45 \\
    Search-o1-7B & 0.47 & 0.13 & 0.16 & 0.05 & 0.10 & 0.14 & 0.20 & 0.18 & 0.21 & 0.36 & 0.42 \\
    Search-R1-7B & 0.59 & \underline{0.54} & \underline{0.61} & \textbf{0.12} & \textbf{0.20} & \textbf{0.40} & \textbf{0.54} & \underline{0.35} & \underline{0.39} & \underline{0.58} & \underline{0.67} \\

    \midrule
    \multicolumn{12}{l}{\textbf{\textit{Interleaved reasoning and retrieval (retriever-free)}}} \\
    \rowcolor[HTML]{C6F2FF}
    \textsc{Freeson-7B} (Ours) & \textbf{0.71} & \textbf{0.55} & \textbf{0.63} & \underline{0.11} & \textbf{0.20} & \underline{0.33} & \underline{0.44} & \textbf{0.39} & \textbf{0.43} & 0.55 & 0.64 \\

    \bottomrule
\end{tabular}
}
\caption{Overall performance on single- and multi-hop QA, and fact-checking tasks. \textbf{Bold} and \underline{underlining} indicate the best and second-best results, respectively. FREESON operates without an external retriever, yet matches or outperforms the baselines.}
\label{table:main_table}
\end{table*}

\begin{table*}[t!]
\centering
\fontsize{8}{11}\selectfont
\begin{tabular}{lccc}
    \toprule
    \multirow{2}{*}{\textbf{Method}} & \textbf{MedQA-US} & \textbf{PubMedQA} & \textbf{BioASQ} \\
    & ACC & ACC & ACC \\
    \midrule
    \multicolumn{4}{l}{\textbf{\textit{Medical LMs}}} \\
    MedGemma-27B & \textbf{0.88} & 0.43 & \textbf{0.81} \\
    BioMistral-7B & 0.39 & 0.19 & 0.73 \\
    MedGemma-1.5-4B & 0.67 & 0.51 & 0.67 \\

    \midrule
    \multicolumn{4}{l}{\textbf{\textit{Retrieve-then-reason}}} \\
    \multicolumn{4}{l}{GPT-3.5 (Xiong et al.~\citep{xiong-etal-2024-benchmarking})} \\
    + BM25       & 0.67 & 0.30 & 0.60 \\
    + Contriever & 0.67 & 0.27 & 0.60 \\
    + SPECTER    & 0.67 & 0.26 & 0.56 \\
    + MedCPT     & 0.66 & 0.29 & 0.61 \\
    + RRF-2      & 0.66 & 0.31 & 0.59 \\
    + RRF-4      & 0.66 & 0.30 & 0.61 \\

    \midrule
    \multicolumn{4}{l}{\textbf{\textit{Interleaved reasoning and retrieval (with retriever)}}} \\
    Gemma-3-27B + MedCPT & 0.80 & 0.24 & 0.78 \\
    \multicolumn{4}{l}{Search-R1-7B} \\
    + E5 & 0.57 & 0.37 & 0.68 \\
    + MedCPT & 0.60 & 0.32 & 0.71 \\

    \midrule
    \multicolumn{4}{l}{\textbf{\textit{Interleaved reasoning and retrieval (retriever-free)}}} \\
    \rowcolor[HTML]{C6F2FF}
    \textsc{Freeson-7B} (Ours) & 0.63 & \textbf{0.56} & \textbf{0.81} \\
    \bottomrule
\end{tabular}
\caption{Zero-shot transfer to the medical domain. FREESON is evaluated using the same experimental settings as in Table~\ref{table:main_table}, replacing only the corpus with 125.8K medical textbook snippets. FREESON-7B achieves competitive performance with MedGemma-27B.}
\label{table:medical_qa}
\end{table*}

\textbf{Medical QA.} For reasoning without retrieval, we evaluate three medical LMs: \textsc{MedGemma-27B}, \textsc{MedGemma-1.5-4B}~\citep{Sellergren_2026}, and \textsc{BioMistral-7B}~\citep{labrak2024biomistralcollectionopensourcepretrained}. For retrieve-then-reason, we report \textsc{GPT-3.5} results with \textsc{BM25}, \textsc{Contriever}, \textsc{SPECTER}, \textsc{MedCPT}, \textsc{RRF-2}, and \textsc{RRF-4} from Xiong et al.~\citep{xiong-etal-2024-benchmarking}. RRF (Reciprocal Rank Fusion) combines retrieval rankings, with \textsc{RRF-2} fusing BM25 and MedCPT, and \textsc{RRF-4} combining all four retrievers. For interleaved reasoning and retrieval, we evaluate \textsc{Gemma-3-27B} with \textsc{MedCPT}~\citep{Jin_2023}, a medical-domain retriever, and \textsc{Search-R1-7B} with \textsc{E5} or \textsc{MedCPT}. The GPT-3.5 results are reported on the full datasets, while our evaluations use subsets sampled with random seed 42.






\subsection{Main Results}
\label{section:main_results}

Tables~\ref{table:main_table} and~\ref{table:medical_qa} present the performance of \textsc{FREESON} on nine benchmarks covering general-domain QA, fact-checking, and medical QA.

\paragraph{Retriever-free reasoning on general-domain tasks.}
Despite operating without a separate retriever, \textsc{FREESON} achieves the highest accuracy of 0.71 on the fact-checking benchmark FEVER, a 12-point gain over Search-R1. On five general-domain QA benchmarks, it achieves performance comparable to Search-R1, with the best results on 2WikiMultihopQA and PopQA and a tie for the best F1 on MuSiQue. Compared with four reasoning baselines using external retrievers, \textsc{FREESON} improves EM and F1 by 14.9 points on average.

\paragraph{Zero-shot transfer to medical QA.}
We further evaluate \textsc{FREESON} by replacing only the Wikipedia corpus with medical textbook snippets, retaining the value head trained on PopQA. Without additional domain-specific training or a medical retriever, \textsc{FREESON} improves average accuracy by 12.5 points over Search-R1 baselines using E5 or MedCPT. It also achieves competitive performance with the much larger medical language model MedGemma-27B, surpassing it on PubMedQA and matching it on BioASQ.

\section{Analysis}
\subsection{Does FREESON retrieve more precise evidence than fixed top-k retrieval?}
\begin{wraptable}{r}{0.5\textwidth}
\vspace{-0.8em}
\centering
\fontsize{12}{15}\selectfont
\resizebox{\linewidth}{!}{%
\renewcommand{\arraystretch}{1.3}
\begin{tabular}{@{}lcccc@{}}
\toprule
\textbf{Method} & \textbf{Top-k} & \textbf{\# Docs} & \textbf{Coverage} & \textbf{Prec.} \\
\midrule
Search-R1 + E5 & 3 & 2.99 & \textbf{0.87} & 0.34 \\
\textbf{FREESON} & 3 & 1.24 & 0.83 & \textbf{0.81} \\
\bottomrule
\end{tabular}%
}
\caption{Retrieval quality on 2WikiMultihopQA. FREESON achieves 2.4$\times$ higher precision while maintaining comparable evidence coverage.}
\label{table:retrieval_quality}
\vspace{-0.8em}
\end{wraptable}
To assess retrieval quality independently of generation, we measure evidence coverage and precision on 2WikiMultihopQA using gold supporting documents. Coverage measures the fraction of gold documents retrieved across all steps, while precision measures the fraction of retrieved documents that are gold. Search-R1 uses top-k=3, while FREESON retrieves up to three documents. As shown in Table~\ref{table:retrieval_quality}, FREESON maintains comparable coverage (0.83 vs. 0.87) while achieving 2.4$\times$ higher precision (0.81 vs. 0.34), yielding substantially cleaner evidence with fewer documents (1.24 vs. 2.99). Unlike fixed top-k retrieval, FREESON returns only documents containing its explored trajectories, reducing irrelevant context for downstream reasoning.
\begin{table*}[t!]
\centering
\fontsize{9}{11}\selectfont
\begin{tabular}{llcccc}
    \toprule
    \textbf{Model} & \textbf{Value estimation} & \textbf{MedQA-US} & \textbf{PubMedQA} & \textbf{BioASQ} & \textbf{Avg.} \\
    & & ACC & ACC & ACC & ACC \\
    \midrule
    \multirow{3}{*}{GPT-OSS-20B}
    & Self-eval & 85.3 & 56.7 & 84.7 & 75.57 \\
    & vh\_medical & 85.0 & 59.3 & 84.7 & 76.33 \\
    & vh\_general & 85.0 & 58.7 & 85.3 & 76.33 \\
    \midrule
    \multirow{3}{*}{Gemma-4-12B}
    & Self-eval & 85.3 & 54.7 & 79.3 & 73.10 \\
    & vh\_medical & 84.3 & 55.3 & 79.3 & 72.97 \\
    & vh\_general & 87.0 & 56.0 & 80.0 & 74.33 \\
    \midrule
    \multirow{3}{*}{BioMistral-7B}
    & Self-eval & 40.7 & 38.0 & 72.0 & 50.23 \\
    & vh\_medical & 40.7 & 35.3 & 72.0 & 49.33 \\
    & vh\_general & 40.0 & 35.3 & 72.0 & 49.10 \\
    \bottomrule
\end{tabular}
\caption{Comparison of self-evaluation and learned value estimators under in-domain and zero-shot transfer settings. All three approaches achieve comparable downstream QA performance.}
\label{table:medical_value_estimation}
\end{table*}

\subsection{Why CT-MCTS over other decoding strategies?}


\begin{wraptable}{r}{0.5\textwidth}
\vspace{-0.8em}
\centering
\fontsize{9}{12}\selectfont
\resizebox{\linewidth}{!}{%
\begin{tabular}{@{}ccccc@{}}
    \toprule
         & \multicolumn{2}{c}{\textbf{PopQA}} & \multicolumn{2}{c}{\textbf{2WikiMultihopQA}}\\ 
        \cmidrule(lr){2-3} \cmidrule(lr){4-5}
        \textbf{Decoding} & EM & F1 & EM & F1 \\
\midrule
Greedy Search & 0.17 & 0.21 & 0.21 & 0.23 \\
Beam Search   & 0.18 & 0.21 & 0.23 & 0.25 \\
\textbf{CT-MCTS}     & \textbf{0.44} & \textbf{0.45} & \textbf{0.54} & \textbf{0.60} \\
\bottomrule
\end{tabular}
}
\caption{Comparison of decoding strategies.}
\label{table:search_strategy}
\vspace{-0.5em}
\end{wraptable}

Table~\ref{table:search_strategy} compares greedy search, beam search, and CT-MCTS under corpus-constrained decoding, with CT-MCTS consistently outperforming both alternatives. Greedy and beam search are prone to early commitment, as incorrect decoding paths are difficult to recover from. In contrast, CT-MCTS repeatedly explores diverse paths from the root, using value estimates to guide exploration toward answer-bearing regions.




\subsection{Which value estimation strategy works best?}

\label{anlaysis:reward_model_comparison}

\begin{figure}[t!]
{
\centering
    \includegraphics[width=\textwidth]
    {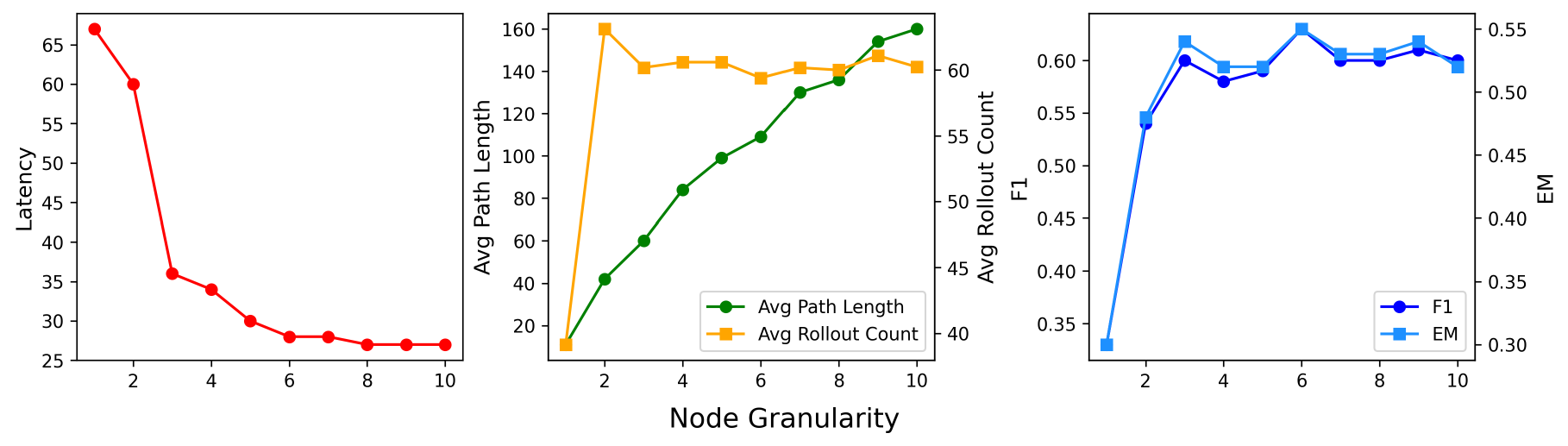}
    \caption{
Impact of node granularity (1 -- 10 tokens per node) on efficiency and performance. \textbf{Left}: Higher granularity reduces latency (s). \textbf{Middle}: Retrieved paths become longer while rollout counts remain similar, indicating more efficient search per computation. \textbf{Right}: Higher granularity generally improves performance.
}
\label{fig:node_granularity}
}
\end{figure}

\begin{table*}[t!]
\centering
    \fontsize{9}{12}\selectfont

    \begin{tabular}{@{}cccccccccc@{}}
        \toprule
             & \multicolumn{2}{c}{\textbf{PopQA}} & \multicolumn{2}{c}{\textbf{TriviaQA}} & \multicolumn{2}{c}{\textbf{2WikiMultiHop}} & \multicolumn{2}{c}{\textbf{Avg.}} \\ 
                \cmidrule(lr){2-3} \cmidrule(lr){4-5} \cmidrule(lr){6-7} \cmidrule(lr){8-9}
               \textbf{Value estimation} & EM & F1
 & EM & F1 & EM & F1 & EM & F1 \\
    \midrule
    $\text{LLM Synthetic Trajectory}$ & 0.41 & 0.41 & 0.48 & 0.61 & 0.53 & 0.59 & 0.47 & 0.54 \\
    $\textbf{CT-MCTS Rollout}$ & \textbf{0.44} & \textbf{0.45} & \textbf{0.51} & \textbf{0.63} & \textbf{0.54} & \textbf{0.60} & \textbf{0.50} & \textbf{0.56} \\
    \bottomrule
    \end{tabular}
\caption{Comparison of on-policy and off-policy value learning. \textsc{CT-MCTS} rollouts yield consistently higher QA performance than LLM-generated synthetic trajectories.}
\label{table:reward_synthetic_real}
\end{table*}

We compare training-free self-evaluation, on-policy, and off-policy value learning. Self-evaluation and on-policy achieve comparable performance, while off-policy performs slightly worse.

\paragraph{On-policy vs. off-policy.}
\label{analysis:value_estimation}
As shown in Table~\ref{table:reward_synthetic_real}, on-policy learning from CT-MCTS rollouts consistently outperforms off-policy learning from LLM-generated trajectories across three QA benchmarks, improving average EM from 0.47 to 0.50 and F1 from 0.54 to 0.56. This suggests that trajectories aligned with inference-time exploration provide more effective training signals.

\paragraph{Self-evaluation vs. learned value estimators.} Table~\ref{table:medical_value_estimation} compares training-free self-evaluation (Self-eval) with value heads trained on medical QA (vh\_medical, in-domain) and PopQA (vh\_general, cross-domain). The medical value heads are trained on CT-MCTS rollouts from 200 MedQA-US and 200 PubMedQA training questions, randomly sampled with seed 42, excluding evaluation samples. Across three reasoning models, all three strategies achieve comparable performance, with no consistent benefit from in-domain training.
This suggests that value estimation primarily involves identifying whether an explored path contains answer-relevant information, rather than requiring specialized knowledge, allowing LRMs to perform retrieval in a fully training-free manner without a separately trained value head.


\subsection{What node granularity is most effective for trajectory exploration?}
\label{analysis:node_granularity}

We investigate multi-token nodes to enable more context-aware decisions in CT-MCTS, balancing semantic richness with fine-grained exploration. As shown in Figure~\ref{fig:node_granularity}, a moderate granularity ($G = 6$) achieves the best performance (F1 $\uparrow$), while reducing latency and yielding longer retrieved paths. Six-token nodes capture coherent semantic units, facilitating more effective exploration than token-level search. At finer granularities ($G = 1$ and $G = 2$), CT-MCTS incurs higher latency and retrieves shorter, less informative paths. The large number of leaf nodes in Table~\ref{table:node_graularity} suggests scattered exploration, potentially requiring more simulations to capture semantic relationships. Conversely, the performance decline from $G = 6$ to $G = 10$ indicates that overly coarse nodes restrict fine-grained exploration. These findings highlight the importance of balancing semantic richness and search controllability through appropriate node granularity.

\subsection{Can LRM-driven multi-node expansion boost performance?}

\begin{wraptable}{r}{0.5\textwidth}
\vspace{-0.8em}
\centering
\fontsize{9}{14}\selectfont
\resizebox{\linewidth}{!}{%
    \begin{tabular}{@{}cccc@{}}
        \toprule
             & \multicolumn{2}{c}{\textbf{Performance}} & \multicolumn{1}{c}{\textbf{Efficiency}} \\
             \cmidrule(lr){2-3} \cmidrule(lr){4-4}
             \textbf{Multiple Expansion} & EM & F1 & Latency \\
        \midrule
        $\text{M=1}$ & 0.41 & 0.46 & \textbf{21s} \\
        $\textbf{M=2}$ & \textbf{0.54} & \textbf{0.60} & 31s \\
        $\text{M=3}$ & 0.54 & 0.60 & 36s \\
        $\text{M=4}$ & 0.54 & 0.60 & 44s \\
        \bottomrule
    \end{tabular}%
}
\caption{Effect of the number of expanded nodes ($M$) per simulation. On 2WikiMultihopQA, performance improves significantly at $M{=}2$ and quickly saturates due to constrained token space.}
\label{table:multiple_expansion}
\vspace{-0.5em}
\end{wraptable}

To better leverage the LRM's reasoning ability over document expressions during expansion in \textsc{CT-MCTS}, we explore expanding multiple candidate child nodes per simulation, selected via multinomial sampling over top-$k$ predictions from the LRM. As shown in Table~\ref{table:multiple_expansion}, setting $M = 2$ achieves the best performance, improving EM and F1 scores by 13 and 14 points, respectively, compared to $M = 1$.

This improvement stems from the increased involvement of the LRM in selecting the node to expand, providing more informed guidance. However, under constrained decoding, the number of valid next-token candidates is inherently limited. As a result, although expanding more candidates is initially beneficial, the performance quickly saturates, as many of the additional nodes are likely to be explored by later simulations anyway. From $M = 3$ onward, no further gains are observed. These findings indicate that even a modest increase in LRM involvement can meaningfully improve retrieval quality in \textsc{CT-MCTS}.
\label{analysis:multiple_expansion}

\subsection{Efficiency of FREESON over embedding-based retrieval}

\begin{wraptable}{r}{0.5\textwidth}
\vspace{-0.8em}
\centering
\fontsize{12}{16}\selectfont
\resizebox{\linewidth}{!}{%
    \begin{tabular}{@{}ccc@{}}
        \toprule
             & \multicolumn{2}{c}{\textbf{Efficiency}} \\
             \cmidrule(lr){2-3}
             \textbf{Metric} & FREESON & Dual-encoder \\
        \midrule
        $\text{Indexing time ($\downarrow$)}$ & \textbf{64 mins} (cpu) & 177 mins (gpu) \\
        $\text{Index memory ($\downarrow$)}$ & \textbf{11GB} & 61GB \\
        $\text{Inference latency ($\downarrow$)}$ & 17.21s & \textbf{9.36s} \\
        $\text{Inference FLOPs ($\downarrow$)}$ & $\sim 5.5 \times 10^{13}$ & $\boldsymbol{\sim 3.1 \times 10^{13}}$ \\
        \bottomrule
    \end{tabular}%
}
\caption{Comparison of indexing time, index memory, inference latency, and FLOPs between FREESON and Search-R1. FREESON builds an FM-index, while the dual encoder uses embedding-based indexing.}
\label{table:efficiency_comparison}
\vspace{-0.5em}
\end{wraptable}
On the 21M Wikipedia corpus, FREESON achieves approximately 3$\times$ faster indexing (63.7 vs. 176.9 mins) and 5.5$\times$ lower index memory (11GB vs. 61GB), even below the 14GB raw corpus size due to BWT compression. Unlike dual encoders, whose retrieval cost grows with the number of indexed documents~\citep{kim2024exploringpracticalitygenerativeretrieval}, FREESON performs corpus-constrained decoding, with computation that does not directly scale with corpus size.

However, CT-MCTS incurs 1.8$\times$ higher inference latency (17.21s vs. 9.36s) and FLOPs ($\sim 5.5 \times 10^{13}$ vs. $\sim 3.1 \times 10^{13}$), reflecting the cost of inference-time exploration. To improve inference efficiency, we (1) skip model inference and directly retrieve subsequent tokens from CorpusTree when only one valid continuation exists, (2) reuse the KV-cache across search branches, (3) cache shared prefixes during value estimation, and (4) reuse the KV-cache across reasoning steps. These optimizations reduce redundant computation without affecting the results.


\section{Conclusion}
We revisit conventional retrieval-augmented reasoning and propose \textsc{FREESON}, a unified framework where a single model serves as both reasoner and retriever, mitigating the representation bottleneck while avoiding the training and maintenance costs of a separate retriever. At its core, CT-MCTS enables LRMs to directly navigate the raw corpus through constrained decoding, without relying on fixed top-k retrieval. Across nine benchmarks, \textsc{FREESON} achieves competitive or improved performance while retrieving more precise evidence with lower indexing costs. Furthermore, training-free self-evaluation performs comparably to learned value estimators without requiring additional value model training. These findings demonstrate that retrieval can be treated as an integral part of reasoning rather than a separate process.

\section{Limitations and Future Work}
A key limitation of the current FREESON framework lies in its relatively simple sub-CorpusTree construction, which may exclude promising search paths. Future work could explore more adaptive strategies for narrowing the search space. Second, CT-MCTS incurs higher inference latency than conventional retrieval methods. Further efficiency improvements could involve selectively activating model parameters needed for retrieval, rather than using the full model at each search step. Finally, CT-MCTS relies on initial entry points and sequential corpus traversal, which may limit exploration efficiency when pieces of evidence are scattered across distant locations within or across documents. More flexible exploration strategies, such as diffusion-inspired exploration, could help locate and integrate such distributed evidence more effectively. Additionally, LRMs could be further trained through reinforcement learning to generate more effective search trajectories during corpus exploration.

\bibliographystyle{unsrt} 
\bibliography{neurips_2025}
\newpage
\appendix
\section{Dataset Statistics \& Retrieval Settings}
\label{appendix:stats}
Table~\ref{table:dataset_stats} presents dataset statistics and retrieval settings for all nine benchmarks. Our pipeline involves iterative retrieval and multiple LLM calls per instance, making full-set evaluation time-consuming. We therefore use benchmark subsets for our main experiments and additionally evaluate on the full 2WikiMultihopQA test set. As shown in Table~\ref{tab:subset_full}, the results exhibit consistent performance trends across both settings.

\begin{table}[ht]
\centering
\fontsize{10}{13}\selectfont

\resizebox{\textwidth}{!}{%
\begin{tabular}{lccccccc}
\toprule
\textbf{Settings} & \textbf{FEVER} & \textbf{PopQA} & \textbf{TriviaQA} & \textbf{HotPotQA} & \textbf{2WikiMultihopQA} & \textbf{MuSiQue} & \textbf{TriviaQA} \\
& \textit{\cite{thorne-etal-2018-fever}} & \textit{\cite{mallen2023trustlanguagemodelsinvestigating}} & \textit{\cite{joshi2017triviaqalargescaledistantly}} & \textit{\cite{yang2018hotpotqadatasetdiverseexplainable}} & \textit{\cite{ho-etal-2020-constructing}} & \textit{\cite{trivedi-etal-2022-musique}} & \textit{\cite{joshi2017triviaqalargescaledistantly}} \\
\midrule
\multicolumn{7}{l}{\textbf{\textit{Dataset statistics}}} \\
\# Examples & 100 & 500 & 500 & 500 & 500 & 500 & 500 \\
Gold answer count & Single & Many & Single & Single & Single & Single & Single \\
\midrule
\multicolumn{7}{l}{\textbf{\textit{Retrieval settings}}} \\
Corpus & Wikipedia-dpr & Wikipedia-dpr & Wikipedia-dpr & Wikipedia-dpr & Wikipedia-2wiki & Wikipedia-dpr & Wikipedia-dpr \\
Corpus size & 21M & 21M & 21M & 21M & 6M & 21M & 21M \\
Retriever & E5 & E5 & E5 & E5 & E5 & E5 & E5 \\
Top-k & 3 & 2 & 3 & 3 & 3 & 3 & 3 \\
\bottomrule
\end{tabular}%
}

\vspace{12pt}

\resizebox{0.75\textwidth}{!}{%
\begin{tabular}{lccc}
\toprule
\textbf{Settings} & \textbf{MedQA-US} & \textbf{PubMedQA} & \textbf{BioASQ} \\
& \textit{\cite{jin2020diseasedoespatienthave}} & \textit{\cite{jin2019pubmedqadatasetbiomedicalresearch}} & \textit{\cite{krithara2023bioasq}} \\
\midrule
\multicolumn{4}{l}{\textbf{\textit{Dataset statistics}}} \\
\# Examples & 300 & 150 & 150 \\
Gold answer count & Single & Single & Single \\
\midrule
\multicolumn{4}{l}{\textbf{\textit{Retrieval settings}}} \\
Corpus & Medical textbooks & Medical textbooks & Medical textbooks \\
Corpus size & 125.8K & 125.8K & 125.8K \\
Retriever & E5 / MedCPT & E5 / MedCPT & E5 / MedCPT \\
Top-k & 3 & 3 & 3 \\
\bottomrule
\end{tabular}%
}

\vspace{6pt}
\caption{Dataset statistics and retrieval settings for general-domain and medical QA benchmarks. We randomly sample evaluation examples to reduce computational cost. ``Many'' denotes multiple gold answers, while ``Single'' denotes one.}
\label{table:dataset_stats}
\end{table}
\begin{table}[!htbp]
\centering
\fontsize{9}{12}\selectfont
\begin{tabular}{lcc}
\toprule
Method & Subset ($N=500$) & Full Set ($N=12{,}576$) \\
& EM / F1 & EM / F1 \\
\midrule
Search-R1 + BM25 & -- & 0.5394 / 0.6085 \\
Search-R1 + E5 & 0.54 / 0.61 & 0.5558 / 0.6360 \\
\textbf{FREESON (Ours)} & \textbf{0.55 / 0.63} & \textbf{0.5742 / 0.6440} \\
\bottomrule
\end{tabular}
\caption{Comparison of subset and full-set evaluation on 2WikiMultihopQA. The results show consistent performance trends across both settings.}
\label{tab:subset_full}
\end{table}
\clearpage
\begin{table*}
\centering
    \fontsize{10}{15}\selectfont
\resizebox{0.93\textwidth}{!}{

    \begin{tabular}{@{}ccccccccc@{}}
        \toprule
             & \multicolumn{3}{c}{\textbf{Efficiency}} & \multicolumn{2}{c}{\textbf{Performance}} & \multicolumn{1}{c}{\textbf{Exploration}} \\ 
                \cmidrule(lr){2-4} \cmidrule(lr){5-6} \cmidrule(lr){7-7}  
               \textbf{Node Granularity} & \textbf{Latency} ($\downarrow$) & \textbf{Avg Rollout Count} & \textbf{Avg Path Length.} & \textbf{EM} ($\uparrow$) & \textbf{F1} ($\uparrow$) & \textbf{Exploration} \\
    \midrule
    $\text{G=1}$ & 67s & 39.14 & 11 & 0.30 & 0.33 & 32 \\
    $\text{G=2}$ & 60s & 63.13 & 42 & 0.48 & 0.54 & 16 \\
    $\text{G=3}$ & 36s & 60.19 & 60 & 0.54 & 0.60 & 12 \\
    $\text{G=4}$ & 34s & 60.60 & 84 & 0.52 & 0.58 & 10 \\
    $\text{G=5}$ & 30s & 60.60 & 99 & 0.52 & 0.59 & 9 \\
    $\cellcolor[HTML]{E0FFFF} \text{G=6}$ & \textbf{28s} & 59.40 & 109 & \textbf{0.55} & \textbf{0.63} & 9 \\
    $\text{G=7}$ & 28s & 60.18 & 130 & 0.53 & 0.60 & 9 \\
    $\text{G=8}$ & 27s & 60.00 & 136 & 0.53 & 0.60 & 8 \\
    $\text{G=9}$ & 27s & 61.10 & 154 & 0.54 & 0.61 & 9 \\
    $\text{G=10}$ & 27s & 60.23 & 160 & 0.52 & 0.60 & 8 \\
    \bottomrule
    \end{tabular}
    }
\caption{Analysis of efficiency, performance, and exploration behavior under varying node granularity. \* Avg Rollout Counts for $G=4$ and $G=5$ are imputed with the mean (excluding $G=1$), as single-token nodes ($G=1$) often fail to yield valid rollouts due to limited semantic context.}
\label{table:node_graularity}
\end{table*}

\section{Analysis on Varying Node Granularity}
Table~\ref{table:node_graularity} shows the efficiency, performance, and exploration behavior under varying node granularity. As granularity increases, latency decreases, while the average path length grows—indicating that more informative reasoning paths are explored. Performance also improves with increasing granularity, peaking at $G=6$ and saturating thereafter. Interestingly, although the number of valid rollouts required during the CT-MCTS process remains theoretically similar or even lower, the length of the retrieved paths increases with granularity. This is because longer current nodes provide more contextual constraints, reducing the number of valid tokens available for rollout at the next step. In other words, increased granularity allows the model to extract more informative reasoning paths with comparable or even reduced computational cost. In the case of $G=1$, \textsc{FREESON} must not only search for answer-containing paths but also infer semantic relationships between adjacent nodes, leading to broader exploration across many nodes. A higher number of explorations in this setting should not be interpreted as a positive signal.

\section{Limitations of Embedding-Based Ground-Truth Annotations in QA Benchmarks}
We observe cases in PopQA and HotpotQA where both Search-R1 + E5 and \textsc{FREESON} (ours) produce semantically correct answers, but only the former matches the ground-truth annotations due to differences in wording. For example, Search-R1 + E5 outputs ``water'' or ``music'', while \textsc{FREESON} produces more specific alternatives such as ``fluid'' or ``opera'', which are not recognized as correct under exact-match evaluation.

Closer inspection reveals that the ground-truth answers often align with the phrasing in documents retrieved by E5. This suggests a potential bias in the ground-truth annotations toward expressions favored by embedding-based retrieval, which may inadvertently advantage such systems during evaluation. We provide examples from HotpotQA below.

\label{appendix:errorcase_hotpot}

\begin{errorbox}[\#1. FREESON retrieves correct documents but ground-truth fails to recognize valid answers]
\begin{lstlisting}[breaklines=true, breakindent=0pt, basicstyle=\ttfamily\small]
-----------------------------------------------------------------
*Question: What substance is always in a trap to prevent the passage of anything from either direction,  even sewer gases? 
-----------------------------------------------------------------
*Ground-truth answer: water
-----------------------------------------------------------------

<think>I need to find the substance that is always in a trap to prevent the passage of anything from either direction, even sewer gases. I'll search for it.</think>


<search> (subject : trap, question : What substance is always in a trap to prevent the passage of anything from either direction, even sewer gases) </search>


<information>For trap : Doc 0. trap (plumbing) trap (plumbing) in plumbing, a trap is a device shaped with a bending pipe path to retain fluid to prevent sewer gases from entering buildings while allowing waste to pass through. in oil refineries, traps are used to prevent hydrocarbons and other dangerous gases from escaping through drains. ... (omitted) ... 

Doc 1. trap (plumbing) venting. certain drum-styled traps are also discouraged or banned. trap (plumbing) in plumbing, a trap is a device shaped with a bending pipe path to retain fluid to prevent sewer gases from entering buildings while allowing waste to pass through. ... (omitted) ...  </information>


<think>I found out that in plumbing, a trap is a device shaped with a bending pipe path to retain fluid to prevent sewer gases from entering buildings while allowing waste to pass through. Now I need to find the substance that is always in a trap to prevent the passage of anything from either direction.</think>

... (omitted) ... 

-----------------------------------------------------------------
<answer> fluid </answer>
-----------------------------------------------------------------
  \end{lstlisting}
\end{errorbox}

\begin{errorbox}[\#2. FREESON retrieves correct documents but ground-truth fails to recognize valid answers]
\begin{lstlisting}[breaklines=true, breakindent=0pt, basicstyle=\ttfamily\small]
-----------------------------------------------------------------
*Question: What type of art does The Consul and Arlecchino have in common?
-----------------------------------------------------------------
*Ground-truth answer: music
-----------------------------------------------------------------

#First step
<think>I need to find the type of art that The Consul and Arlecchino have in common. I'll search for it.</think>

<search> (subject: The Consul, question: What type of art does The Consul have) </search>

<information>For The Consul : Doc 0. the consul the consul the consul is an opera in three acts with music and libretto by gian carlo menotti, his first full-length opera. ... (omitted) ...

Doc 1. the consul prize for music and also the 1950 new york drama critics' circle award for best musical. ... (omitted) ...

Doc 2. the consul the broadway cast, this time with rosemary kuhlmann as the secretary of the consulate. ... (omitted) ... noted the topicality of the opera by analogy to the real-life situations </information>


#Second step
<think>I found out that The Consul is an opera. Now I need to find the type of art that Arlecchino has in common with The Consul.</think>

<search> (subject: Arlecchino, question: What type of art does Arlecchino have) </search>

<information>For Arlecchino : Doc 0. arlecchino (opera) of a double-bill. the first british staged performance of arlecchino was in 1954 at glyndebourne. however, edward clark had produced a concert version in london in 1939. the opera is in four movements ... (omitted) ...

Doc 1. arlecchino (opera) arlecchino, oder die fenster (harlequin, or the windows, is a one-act opera with spoken dialog by ferruccio busoni. ... (omitted) ...

Doc 2. arlecchino (opera) (finale.) in procession the other characters, leandro and colombina, dottore and abbate, the donkey and the cart, the two constables, and finally arlecchino and annunziata, cross the stage and bow to the audience ... (omitted) ... </information>

<think>I found out that Arlecchino is an opera. Now I can compare the type of art that The Consul and Arlecchino have in common. Both The Consul and Arlecchino are operas.</think>

-----------------------------------------------------------------
<answer> opera </answer>
-----------------------------------------------------------------
  \end{lstlisting}
\end{errorbox}

\section{Prompt Template for \textsc{FREESON}}\label{appendix:prompt_template}

\subsection{Reasoning prompt template}\label{appendix:r2r_reasoning_prompt}

\begin{instructionsbox}[Reasoning process]
\begin{lstlisting}[breaklines=true, breakindent=0pt, basicstyle=\ttfamily\small]
Answer the given question.
You must conduct reasoning inside <think> and </think> every time you get new information.
After reasoning, if you find you lack some knowledge, you can call a search engine by:
<search> (subject : Help! Help! Police!, question : Who is the director of the film Help! Help! Police!) </search>
This is the correct form for the query: 'Who is the director of the film Help! Help! Police?'
It will return the searched results between <information> and </information>.
You can search as many times as you want.
If you find no further external knowledge is needed, you can directly provide the answer inside <answer> and </answer>,
without detailed illustrations. For example: <answer> Beijing </answer>
Only respond to the final question. Your answer must reflect the end goal, not just a part of the process.
Question: {question}
\end{lstlisting}
\end{instructionsbox}

\subsection{Retrieval prompt template}\label{appendix:r2r_retrieval_prompt}

\begin{instructionsbox}[Retrieval prompts for reasoning over document structure]
\begin{lstlisting}[breaklines=true, breakindent=0pt, basicstyle=\ttfamily\small]
Given a subject and a question, generate a word or phrase likely to appear in a document 
that answers the question.

Q: subject: Star Wars, question: who did Star Wars direct?
A: Star Wars is directed by

Q: subject: Alice, question: When was Alice born?
A: Alice (January 1, 1970 ~ December 12, 2024)

Now your turn:
Q: {question}
A: 
\end{lstlisting}
\end{instructionsbox}
\raggedbottom
\subsection{Prompt template for training-free self-evaluation}
\label{appendix:self_evaluation_prompt}

\begin{instructionsbox}[Training-free self-evaluation]
\begin{lstlisting}[breaklines=true, breakindent=0pt, basicstyle=\ttfamily\small]
You are judging a text segment produced by a retriever over  medical textbooks.

Query: {query}
Segment: {segment}

Is this segment answer-bearing? A segment is answer-bearing if it states, or clearly comes from a passage that states, at least part of the answer to the query, so that reading the passage it comes from would let you answer the query.
A segment that only mentions the subject without pointing to the answer is NOT answer-bearing.
Reply with only Yes or No.
\end{lstlisting}
\end{instructionsbox}
The score is computed as the normalized probability of ``Yes'', yielding a continuous score between 0 and 1.

\subsection{CT-MCTS value network prompt template for training and inference}\label{appendix:r2r_reward_prompt}

\begin{instructionsbox}[Training on-policy value networks and evaluation using them]
\begin{lstlisting}[breaklines=true, breakindent=0pt, basicstyle=\ttfamily\small]
# training
Answer the given question. 
you can call a search engine using <search> and </search>. 
It will return the top searched results between <information> and </information>. 
Based on the provided information, provide the final answer inside <answer> and </answer>.
Question: {question}


# inference
Score from 0 to 1 how much the generated reference contains at least a partial answer to the query.
Query: {query_text}
Generated reference: {rollout_text}
Score:
\end{lstlisting}
\end{instructionsbox}

\subsection{Prompt template for training value networks with synthetic rollouts}
\label{appendix:value_network_synthetic}

\begin{instructionsbox}[Training off-policy value networks]
\begin{lstlisting}[breaklines=true, breakindent=0pt, basicstyle=\ttfamily\small]

    You are helping build a dataset for a reward model.\n\n
    Given:\n
    - A user query\n
    - A reference sentence that correctly answers it\n\n
    Your task:\n
    1. Generate 3 diverse outputs that vary in:\n
       - Whether they contain the exact answer\n
       - Helpfulness in answering the query\n
       - Length and form (sentence, phrase, or word)\n
    2. Include at least 1 short or fragment-style response.\n\n
    Each output should be a dictionary with:\n
    - 'generated': the output\n
    - 'has_answer_score': 1 only if it contains the exact answer textually (not paraphrased)\n
    - 'sim_seq_score': float (0.0-1.0) based on how well it answers the query\n\n
    Example:\n
    Query: What nationality is Aleksandr Stolper?\n
    Reference: Aleksandr Borisovich Stolper (12 August 1907, Dvinsk (now Daugavpils) - 12 January 1979, Moscow) was a Russian/Soviet film director and screenwriter.\n
    Output: [{{\"generated\": \"Aleksandr Borisovich Stolper\", \"has_answer_score\": 0, \"sim_seq_score\": 0.7}}, 
    {{\"generated\": \"Aleksandr Borisovich Stolper (12 August 1907, Dvinsk (now Daugavpils))\", \"has_answer_score\": 0, \"sim_seq_score\": 0.85}}, 
    {{\"generated\": \"Russian/Soviet film director and screenwriter.\", \"has_answer_score\": 1, \"sim_seq_score\": 0.7}}]\n\n
    Now do the same for:\n
    Query:\n{query}\n\nReference:\n{reference}\n\nOutput:
    \end{lstlisting}
\end{instructionsbox}

\section{Evaluation Details}\label{appendix:eval_details}

For general-domain QA, the retrieval-based baselines use Qwen2.5-7B. The FREESON framework performs retriever-free inference based on the SearchR1-nq\_hotpotqa\_train-qwen2.5-7b-em-ppo checkpoint from Search-R1~\citep{jin2025searchr1trainingllmsreason}.

Hyperparameters for all models are set as follows:
\begin{itemize}
    \item Temperature: 0.7
    \item Max new tokens per reasoning step: 1,024
\end{itemize}

For <search> steps in FREESON:
\begin{itemize}
    \item Max simulation: 20
    \item Max rollout length: 30
\end{itemize}

\section{Use of Large Language Models}
LLMs were used to assist with language polishing and the implementation of some baselines.

\end{document}